\documentclass[11pt]{article}
\usepackage{cereb}
\setcitestyle{numbers,square,citesep={,},sort&compress}
\usepackage[utf8]{inputenc}
\usepackage{amsmath,amssymb,graphicx,microtype}
\usepackage{booktabs,multirow,array,enumitem}
\usepackage{algorithm,algpseudocode,wrapfig,tabularray,pifont}
\usepackage{xcolor}
\usepackage[normalem]{ulem}
\usepackage{hyperref}
\hypersetup{hidelinks,pdfauthor={}}
\usepackage{docmute}
\usepackage{etoolbox}
\usepackage{dneo_cerebras_extras}
\graphicspath{{./}{../}{figures/}{images/}{assets/}
  {dneo_iclr2027_project/}{dneo_iclr2027_project/figures/}
  {dneo_iclr2027_project/images/}{dneo_iclr2027_project/assets/}}

\newcommand{\paperfigure}[2]{\includegraphics[width=#1]{#2}}
\definecolor{tickolive}{RGB}{107,142,35}
\definecolor{crossred}{RGB}{190,55,55}
\providecolor{olive}{RGB}{107,142,35}
\providecolor{teal}{RGB}{0,128,128}
\newcommand{\yesmark}{\textcolor{tickolive}{\ding{51}}}
\newcommand{\nomark}{\textcolor{crossred}{\ding{55}}}
\providecommand{\reviewcomment}[2]{}
\providecommand{\reviewdel}[1]{}

\let\DNeoOriginalBibStyle\bibliographystyle
\renewcommand{\bibliographystyle}[1]{\DNeoOriginalBibStyle{cereb}}
\let\DNeoOriginalBib\bibliography
\renewcommand{\bibliography}[1]{%
  \IfFileExists{#1.bib}{\DNeoOriginalBib{#1}}{%
    \IfFileExists{dneo_iclr2027_project/#1.bib}{%
      \DNeoOriginalBib{dneo_iclr2027_project/#1}}{\DNeoOriginalBib{#1}}}}

\title{DNeo: Your Parallel Draft Model Needs Better Target Feature Encoding}
\author{
Jameson Sandler\textsuperscript{*,1},
Shane Bergsma\textsuperscript{2},
Mostafa Elhoushi\textsuperscript{2},
Bin Claire Zhang\textsuperscript{2}, \\
Gavia Gray\textsuperscript{2},
Joel Hestness\textsuperscript{2},
Nolan Dey\textsuperscript{$\dagger$,2},
Ferdinando Fioretto\textsuperscript{$\dagger$,1}
}         
\AfterEndEnvironment{abstract}{\DNeoResourceLinks}

\renewenvironment{quote}{\begin{dneofinding}}{\end{dneofinding}}
\let\DNeoOriginalItalic\textit
\makeatletter
\RenewDocumentCommand{\textit}{m}{%
  \begingroup\edef\dneo@probe{\detokenize{#1}}%
  \ifnum\pdfstrcmp{\dneo@probe}{Rejected proposals are a valuable source of context, which the drafter exploits best by attending to them rather than fusing them into its inputs.}=0\relax
    \endgroup\begin{dneofinding}\DNeoOriginalItalic{#1}\end{dneofinding}%
  \else
    \ifnum\pdfstrcmp{\dneo@probe}{The DNeo protocol encodes more useful target layers, taking advantage of additional target layers.}=0\relax
      \endgroup\begin{dneofinding}\DNeoOriginalItalic{#1}\end{dneofinding}%
    \else
      \endgroup\DNeoOriginalItalic{#1}%
    \fi
  \fi}
\makeatother

\begin{document}
\maketitle
\vspace{-5pt}
\begin{abstract}
Speculative-diffusion decoding uses a small parallel \textit{drafter} model to generate $b$ \textit{draft} tokens in one forward pass, which then undergo parallel \textit{verification} via a larger \textit{target} model. Faster inference can be achieved when multiple draft tokens pass verification simultaneously. Recent parallel draft architectures ({DFlash}, {DSpark}) improve draft quality by conditioning on rich features computed by the target, through a mechanism known as \textit{feature encoding}.
However, standard feature encoding mechanisms have three limitations: (1) they are \textit{lossy}, applying encoding projections that are {destructive} to the underlying target feature-space via a discovered effect known as \textit{residual mixing}, (2) they are \textit{memory-less}, with no record of previously processed proposals, and (3) they utilize features {sampled uniformly} from across the target layers, omitting important hidden states.
To address these issues, we propose Mixture-of-Features Attention (MoFA), a lightweight feature-encoding approach equipped with a working memory that encodes previously rejected proposals. Leveraging MoFA, rejection memory, and a set of further enhancements, including \textit{2-tap causal convolutions, prefix-only cross-attention, and non-uniform target layer selection}, the final architecture, \textbf{DNeo}, achieves $1.3 \times$ faster generation speed than SoTA (DFlash-2) without impacting drafter footprint or generation quality. DNeo marks a new state of the art in high quality, fast LLM inference.
\end{abstract}
\vspace{+5pt}
\begin{figure}[h]
    \centering    \includegraphics[width=0.75\linewidth]{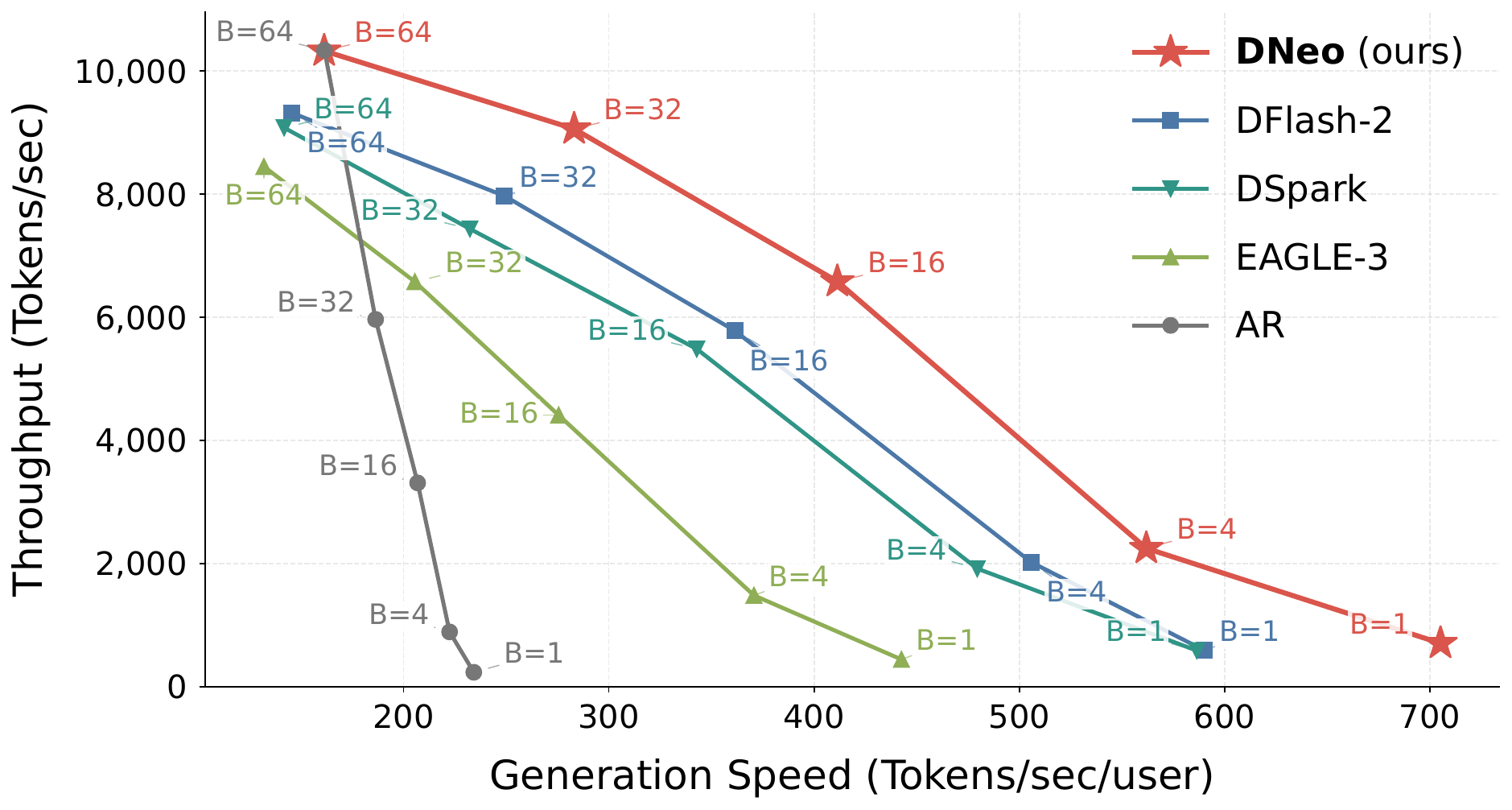}
    \vspace{-12pt}  
    \caption{H100 Inference speeds, with \textbf{Qwen3-4B} ($b=15$) across six benchmarks (see Section \ref{sec:edesign}), and batch-sizes $B$. DNeo dominates the speed Pareto frontier relative to autoregressive (EAGLE-3) algorithms and SoTA parallel algorithms (i.e., DFlash, DFlash-2, DSpark).}
    \label{fig:motivation}
\end{figure}

\section{Introduction}
\label{sec:introduction}
\vspace{-10pt}
Large language models (LLMs) generate text autoregressively, requiring a full forward pass of the model for every generated token. While this serial dependency ensures high-quality text generation, it also imposes a significant computational bottleneck, and poorly exploits the capacity of modern hardware, as illustrated in recent studies on LLM inference efficiency \citep{shazeer2019fast, pope2023efficiently, kwon2023efficient, gholami2024ai}.
Speculative decoding \citep{leviathan2023fast, xia2024unlocking} addresses this serial bottleneck by configuring the LLM to act as a target model that verifies, in parallel, a sequence of token proposals  (a draft) submitted by a smaller drafter model. The verification process applies rejection sampling to the draft to ensure that the target model distribution is respected, guaranteeing no quality loss \citep{leviathan2023fast}. Since verification is parallelized over the draft, speculative decoding can unlock faster generation speeds.

The generation speed of speculative decoding is reduced by drafting latency and is increased by the tokens per forward (TPF), denoting the expected number of tokens committed per speculative round. TPF is determined by draft quality: the number of draft tokens that align with the target model's distribution. Larger drafters can improve draft quality, but often introduce additional latency that offsets these gains \citep{li2026eagle}. Parallel drafters \citep{sandler2026specdiff, li2026diffuspec} achieve low draft latency by generating the entire draft in a single forward pass. However, modeling the dependencies between draft tokens generated in parallel is challenging for drafters, impairing draft quality. To support draft quality, parallel drafters tend to rely on two key mechanisms: \textit{feature encoding}, which conditions the drafter on encoded hidden states of the target model \citep{chen2026dflash, zhang2026dflare}, and \textit{causal heads}, which apply lightweight autoregressive adjustments to the parallel draft posteriors \citep{cheng2026DSpark,inco2026dflash2}.

However, this work reveals flaws in both mechanisms. Feature encoding in state-of-the-art parallel drafters suffers from three limitations: \textit{(i) Residual mixing:} Standard feature encoding \citep{chen2026dflash, inco2026dflash2} uses a dense projection matrix that mixes representations across the hidden state dimension of the target model, an effect that this paper shows produces less useful contextual features. \textit{(ii) Memory-less encoding:} Existing encoders only encode the drafter's context, ignoring the useful features present within previously rejected proposals. \textit{(iii) Uniform layer selection:} Existing encoders heuristically select hidden states uniformly throughout the depth of the target model, without regard to which layers are the most useful for drafting \citep{zhang2026dflare}. Existing causal heads are also limited: they either lack expressivity \citep{cheng2026DSpark} or cannot make global vocabulary corrections \citep{inco2026dflash2}.

To address these limitations, this paper introduces \textbf{DNeo}, a parallel draft architecture that pairs structure-preserving feature encoding with expressive causal adjustments. Its architectural components are designed to enhance LLM generation speed by improving the expected TPF.

\textbf{Contributions.} The paper makes three contributions:
\begin{enumerate}[leftmargin=*,nosep]
    \item It shows that existing feature encoding suffers from residual mixing (Section \ref{sec:mofa_res}), and replaces the dense encoder with Mixture-of-Features Attention (MoFA), a lightweight mixture of target layers that improves TPF by $+0.6$ over dense encoding (Sections \ref{sec:mofa} and \ref{sec:tpf}). It also shows that encoding uniformly spaced hidden states harms draft quality, and introduces the \textit{DNeo protocol}, which isolates the contextually richer later half of the target model (Section \ref{sec:feature-protocol}). 

    \item It shows that previously rejected draft tokens contain useful features, and introduces \textit{rejection memory}, which equips the drafter with cross-attention to these features and improves TPF by $+0.17$ over a memory-less drafter (Section \ref{sec:rejection-memory}). It also introduces the \textit{fused DNeo causal head}, which combines two common causal heads \citep{cheng2026DSpark, inco2026dflash2} in latent space and improves draft quality over both heads in isolation (Section \ref{sec:causal-head}), and streamlines the drafter with prefix-only attention (Section \ref{sec:prefix-only}).

    \item Through extensive experiments in realistic serving settings with SGLang, it shows that DNeo's higher draft quality translates into faster generation than state-of-the-art parallel drafters on common benchmarks outside its training distribution (Section \ref{sec:results}). On chain-of-thought workloads \citep{openai2026openaio1card} with Qwen3-4B \citep{yang2025qwen3technicalreport}, DNeo generates up to $30\%$ faster than DFlash-2 \citep{inco2026dflash2} and delivers $16\%$ higher throughput at batch size 32 (Fig. \ref{fig:motivation}).
\end{enumerate}

\section{Background}\label{sec:background}
This section reviews the two foundations of DNeo: the verification process of speculative decoding, which ensures that generation quality does not degrade, and parallel drafters, which enable low-latency draft generation.

\begin{figure}[t]
    \centering
    \includegraphics[width=0.999\linewidth]{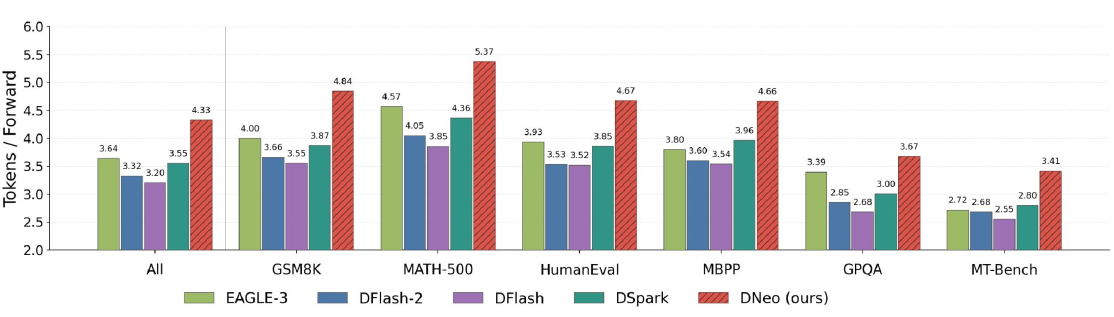}
    \caption{\textbf{Draft quality comparison}. Tokens per forward ($\tau$) over the test-set (Section \ref{sec:edesign}), using official SGLang inference implementations. DNeo dominates existing methods in draft quality.}
    \label{fig:tpfplot}
\end{figure}

\textbf{Speculative Decoding.} 
Autoregressive (AR) language models generate a single token per forward pass, making inference latency directly proportional to output length. Speculative decoding \citep{leviathan2023fast, xia2024unlocking} can accelerate AR inference by replacing serial generation with a parallel \textit{verification} phase. At each speculative round, the smaller \textit{draft} model $q(\cdot)$ proposes a draft of $b$ draft tokens  $(\hat x_1, \dots, \hat x_b) \sim q(\hat x_1, \cdot), \dots, q(\hat x_b, \cdot),$ to be \textit{verified} by the target model $p(\cdot)$ in a single parallel forward pass $p(\hat x_1, \dots, \hat x_b)$, akin to the prefill of a prompt. 

A token $\hat x_i$ from the draft is \textit{accepted}, i.e., committed to the final output, with probability $\Pr[\textit{accept}\; \hat x_i]\triangleq\min(1,  p(\hat x_i, \cdot)/  q(\hat x_i, \cdot)),$ which intuitively scales in proportion to the quality of the proposed draft token $\hat x_i$.
The first rejection at position $\tau$ discards all subsequent tokens $(\hat x_{\tau+1},\dots, \hat x_{b})$ regardless of their quality, and $\hat x_\tau$ is replaced with a \textit{bonus token} $x_\text{bonus}$ sampled from the \textit{correction distribution} $x_\text{bonus} \sim \text{norm}(\max(p(\cdot)-q(\hat x_\tau),0)).$
This criterion ensures that the committed block $(\hat x_1, \dots, \hat x_{\tau-1},x_\text{bonus})$ is theoretically sampled from the target posterior $p(\cdot)$, making decoding \textit{lossless} \citep{leviathan2023fast}.

Each speculative round thus commits $\tau$ tokens, at the cost of the drafting time $t_\text{draft}$ and the verification time $t_\text{verify}$; we refer to the expected $\tau$ as the \textit{tokens per forward} (TPF), and Fig. \ref{fig:tpfplot} compares it across drafters. The average latency per generated token is then, 
\begin{equation*}    \frac{\textit{time}}{\textit{token}}(s)=\frac{t_\text{draft}+t_\text{verify}}{\tau},
\end{equation*}
meaning that latency per token is reduced by increasing $\tau$ through improved draft quality, as long as the speculative overhead $t_\text{draft}+t_\text{verify}$ does not proportionally increase. Raising $\tau$ without proportionally raising this overhead is the goal of every component of DNeo (Section \ref{sec:architecture}).

\textbf{Parallel Token Prediction.}
Parallel token prediction has been an area of significant research interest \citep{diffusiongemmateam2026diffusiongemmatechnicalreport, fu2026nemotronlabsdiffusiontrimodelanguagemodel, yu2026introspectivediffusionlanguagemodels} 
for its potential to reduce LLM inference latency, and support fast speculative drafting \citep{christopher2025speculative, sandler2026specdiff}. The most common form of parallel drafters, masked LMs \citep{sahoo2024simple}, take as input a block of \textit{masked} token embeddings $[\mathbf{m}]_1, \dots, [\mathbf{m}]_b$, the \textit{masked block}, and generate a set of token predictions: 
\begin{equation}
    (\hat{x}_1, \dots, \hat{x}_b) \sim
    q^{\text{parallel}}\left(
    \underbrace{x_1, \dots, x_n}_{\text{prefix}},
    \;\;
    \underbrace{[\mathbf{m}]_1, \dots, [\mathbf{m}]_b}_{\text{masked block}}
    \right)
\end{equation}
from the contextualized hidden state at each masked position. Typically, parallel drafters attend to the \textit{clean} context $\mathbf{x}_\text{ctx}=(x_1, \dots, x_n)$ of committed tokens, drawn from a vocabulary ${\cal V}$, via cross-attention, and perform bidirectional self-attention over the learned masked embeddings \citep{chen2026dflash}. During drafting, this attention structure enables contextual information to be absorbed into masked embeddings, while self-attention is designed to support self consistent draft blocks \citep{chen2026dflash}; though Section \ref{sec:prefix-only} revisits this second choice. 

Supplying parallel models with rich contextual representations, and ensuring that tokens generated in parallel ($\hat x_1, \dots, \hat x_b$) respect left-to-right consistency prove to be critical for improving parallel generation quality \citep{cheng2026DSpark, huang2026domino, chen2026dflash}. Architectural components that support these two goals are reviewed in the next section.

\section{Architectural Components of Parallel Drafters}\label{sec:relatedwork}

Recent parallel token drafters, leveraging the framework of speculative decoding, introduce several architectural components (see Table \ref{table:related}), three of which are directly relevant to DNeo: (\textit{i}) target feature encoding, (\textit{ii}) causal heads, and (\textit{iii}) rejection attention~\citep{chen2026dflash, cheng2026DSpark, zhang2026dflare, inco2026dflash2, zhang2026dflowenablingverifierinformation}. Each component addresses a weakness of parallel drafting, and each leaves a limitation that DNeo addresses (Section \ref{sec:architecture}). 

\textbf{Target Feature Encoding.} 
\textit{Feature encoding} extracts rich target model features of the clean context $\mathbf{x}_\text{ctx}=(x_1, \dots, x_n)$, in order to supply the drafter with representations that improve its ability to make high quality predictions \citep{chen2026dflash, inco2026dflash2, cheng2026DSpark, zhang2026dflare}. During prefill and verification, $c$ hidden states $\vec h^{(l_1)}, \dots, \vec h^{(l_c)}$ from a subset of $c$ target layers $l_1, \dots, l_c$ are concatenated\footnote{Circle, $\circ$, {represents concatenation}} and down-projected into the drafter's $d_h$ dimensional hidden space. The standard encoding approach is a \textit{dense} linear encoder $F_c \in {\cal R}^{cd_h\times d_h}$
\begin{equation}
{\vec f}_i = \text{enc}_{\text{dense}}(x_i) \triangleq \text{RMSNorm} [F_c(\vec h_i^{(l_1)} \circ \dots \circ \vec h_i^{(l_c)})],
\end{equation}
producing the \textit{context feature} encoding $\vec f_i$ of the token $x_i \in \mathbf{x}_{\text{ctx}}$.

Context features $(\vec f_1, \dots, \vec f_n)$ are then used by draft models to compute a contextual key-value cache $(K_{\text{ctx}}^j, V_\text{ctx}^j)$ at every draft layer $\ell_j$ of a drafter with $m$ layers,
\begin{equation}
    K^j_\text{ctx} = W_j^K(\vec f_1, \dots, \vec f_n), \; 
   \text{and}\;\;
    V^j_\text{ctx} =W_j^V(\vec f_1, \dots, \vec f_n),
\end{equation}
via the drafter's key-value projections $W_j^K, W_j^V$. During drafting, masked positions absorb rich prefix information via cross-attention to context features (Fig. \ref{fig:dneomask}), producing draft quality improvements ($+40\%$ larger $\tau$; \citep{chen2026dflash}). However, two choices of this design limit the drafter. First, $F_c$ combines every coordinate of the $c$ hidden states into every coordinate of $\vec f_i$, mixing representations across the hidden dimension; we call this effect \textit{residual mixing}, and show that it degrades the encoded features (Section \ref{sec:mofa_res}). Second, the layers $l_1, \dots, l_c$ are spaced uniformly across the target model, regardless of their usefulness for drafting. Sections \ref{sec:mofa} and \ref{sec:feature-protocol} revisit these two choices.

\begin{table}[t]
\centering
\caption{\textbf{Parallel drafter architecture comparison}. DNeo combines effective components from prior work (causal head, 2-tap convolutions) while also utilizing novel architectural innovations (rejection memory, prefix-only attention, non-uniform target layer selection).}
\label{tab:related_work_comparison}
\label{table:related}

\setlength{\tabcolsep}{5pt}
\renewcommand{\arraystretch}{1.35}
\resizebox{\linewidth}{!}{%
\begin{tabular}{@{}lccccccc@{}}
\toprule
\textbf{Method}
& \shortstack{Causal\\Head}
& \shortstack{Rejection\\Attention}
& \shortstack{Rejection\\Memory\\(\textit{ours})}
& \shortstack{Prefix-only\\Attention\\(\textit{ours})}
& \shortstack{2-Tap\\Convolutions}
& \shortstack{Non-uniform\\Target Layer\\Selection (\textit{ours})}
& \shortstack{Structure\\Preserving\\Encoding} \\
\midrule

DFlash
& \nomark & \nomark & \nomark & \nomark
& \nomark & \nomark & \nomark \\

DFlash-2
& \yesmark & \nomark & \nomark & \nomark
& \yesmark & \nomark & \nomark \\

DFlare
& \nomark & \nomark & \nomark & \nomark
& \nomark & \nomark & \yesmark \\

DFlow
& \yesmark & \yesmark & \nomark & \nomark
& \nomark & \nomark & \nomark \\

DSpark
& \yesmark & \nomark & \nomark & \nomark
& \nomark & \nomark & \nomark \\

\midrule
{\textbf{DNeo}} (\textit{ours})
& \yesmark & \yesmark & \yesmark & \yesmark
& \yesmark & \yesmark & \yesmark \\

\bottomrule
\end{tabular}%
}
\end{table}

\textbf{Causal Heads.}
Parallel draft models have been shown to suffer from a phenomenon known as \textit{suffix decay} \citep{cheng2026DSpark}. By default, parallel drafter \textit{backbones} generate the draft block ($\hat x_1, \dots, \hat x_b$) in a single forward pass, meaning that each draft token does not condition on draft tokens sampled in other parts of the block, leading to left-to-right, or causal, incoherence of the draft \citep{cheng2026DSpark, gu2018nonautoregressiveneuralmachinetranslation, huang2022directedacyclictransformernonautoregressive}. Draft tokens further away from the prefix have fewer directly preceding sampled tokens to condition on, causing draft quality to degrade with draft position: this is suffix decay \citep{cheng2026DSpark}. 

\begin{figure}[t]
    \centering
    \paperfigure{0.95\linewidth}{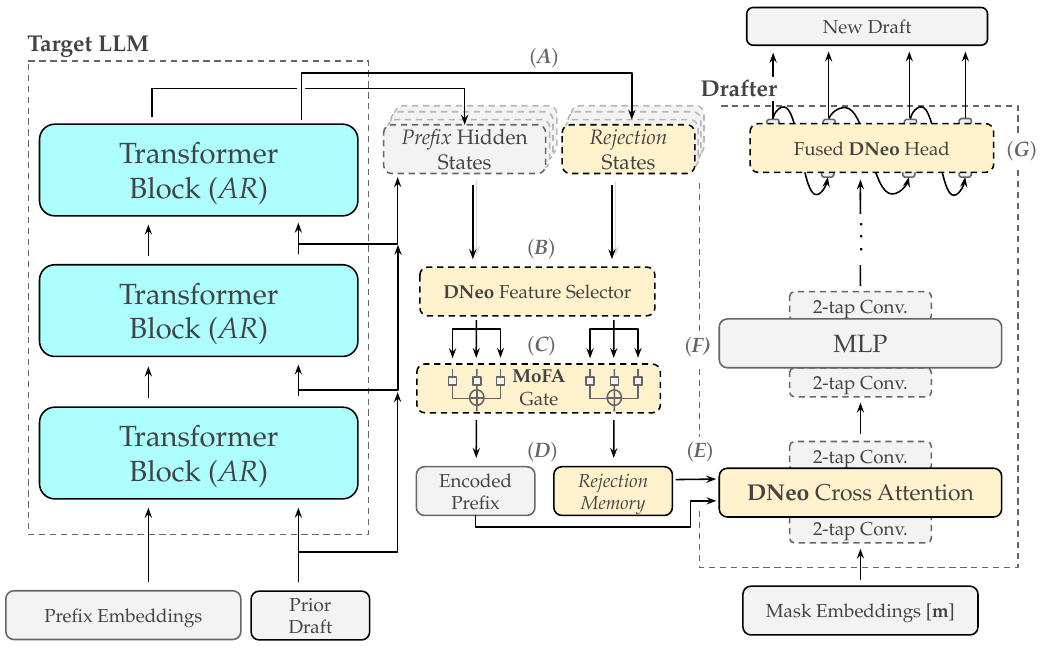}
    \caption{\textbf{DNeo} \textbf{architecture}. Target model (\textit{left}) produces residual streams for the clean-prefix, and rejected positions. Streams are aggregated into \textit{prefix hidden states} and \textit{rejection states} \textit{(A)}. The most useful states are kept with the DNeo feature selector \textit{(B)} (Section \ref{sec:feature-protocol}), then passed to the MoFA gate \textit{(C)} (Section \ref{sec:mofa}) to obtain the \textit{encoded prefix}, and \textit{rejection memory} \textit{(D)} (Section \ref{sec:rejection-memory}). Parallel draft blocks perform cross-attention to the target features \textit{(E)}, utilize 2-tap convolutions \textit{(F)}, and finally apply the fused DNeo head \textit{(G)} (Section \ref{sec:causal-head}) to produce the \textit{new draft}.}
    \label{fig:dneo-architecture}
\end{figure}

Suffix decay in parallel models is generally addressed via a \textit{causal head}, popularized by DSpark \citep{cheng2026DSpark}. Typically, causal heads are parametrized by a low rank matrix $W_r$, 
\begin{equation} \label{eq:dsparkhead}   q^\text{parallel}_{j}(\hat x_j)\triangleq{} \text{softmax}(\vec z_{j} + \underset{\text{Causal Head}}{\underbrace{W_r(\mathbf{1}[\hat x_{j-1}])})},
\end{equation}
which applies a correction to the backbone logits $\vec z_j \in {\cal R}^{|{\cal V}|}$ at each draft position $j$. The adjusted posterior $q^\text{parallel}_{j}(\hat x_j) \in {\cal R}^{|{\cal V} |}$ uses $W_r\in {\cal R}^{|{\cal V}| \times |{\cal V}|}$ of rank $r$, factorized as $W_r=W_\text{down}W_\text{up}$ with $W_\text{down}\in {\cal R}^{|{\cal V}| \times r}$ and $W_\text{up}\in {\cal R}^{r \times |{\cal V}|}$, to condition on the one-hot encoding $\mathbf{1}[\hat x_{j-1}]$ of the previously sampled draft token. The adjusted posterior is computed for all indexes $j \in (1 \dots b)$ in a left-to-right manner, ensuring that each sampled token $\hat x_j$ is causally consistent with its predecessor $\hat x_{j-1}$; this sequential pass is the main cost of a causal head. DFlash-2 \citep{inco2026dflash2} extends this design with a more expressive bilinear score head, which re-scores only the top-$k$ candidates of the backbone (Section \ref{sec:causal-head}), and with \textit{2-tap convolutions}, which mix each masked state with that of its left neighbor before and after every attention and MLP operation (Fig. \ref{fig:dneo-architecture}). However, $W_r$ conditions only on the previous token through a low-rank linear map, while the score head cannot promote tokens outside the top-$k$ candidates; Section \ref{sec:causal-head} combines the strengths of both.

\textbf{Rejection Attention.}
Conventionally, state-of-the-art parallel drafter architectures do not leverage draft information from previous speculative rounds \citep{chen2026dflash, cheng2026DSpark, inco2026dflash2, zhang2026dflare}. This \textit{memory-less} property limits the prediction accuracy of drafters in online settings (Section \ref{sec:rejection-memory}). Concurrent work conditions future rounds on previous rejections, an approach we refer to broadly as \textit{rejection attention} \citep{zhang2026dflowenablingverifierinformation, yu2026dedititerativedraftediting}. For example, the DFlow architecture \citep{zhang2026dflowenablingverifierinformation} computes a feature $\hat f_j= \text{enc}_\text{dense}(\hat x_j)$ of the first rejected token $\hat x_j$, and concatenates it with the token embeddings $W^E[x_\text{bonus}], W^E[\hat x_j] \in {\cal R}^{d_h}$, where $W^E$ is the drafter's embedding matrix. An MLP with parameters $\phi$ processes this concatenation, $\vec v_w, \vec v_b=\text{MLP}_\phi(W^E[ x_{\text{bonus}}]\circ W^E[\hat x_{j}] \circ\hat f_j)$, producing \textit{relay vectors} $\vec v_w, \vec v_b \in {\cal R}^{d_h}$, which gate an update of every masked embedding before the next round\footnote{Via a Hadamard product `$\odot$'}: $[\mathbf{m}]_i \leftarrow [\mathbf{m}]_i + (\vec v_w \odot \hat f_j) + \vec v_b \;\; (\forall i \in 1\ldots b)$. However, DFlow uses only the \textit{first} rejected token, and fuses it into every masked embedding, producing a contextual bottleneck. DNeo builds on rejection attention, introducing \textit{rejection memory} (Section \ref{sec:rejection-memory}), which instead enables masked tokens to attend to \emph{all} rejected tokens.

\section{DNeo Architecture}
\label{sec:architecture}

The outline of the DNeo architecture is shown in Fig. \ref{fig:dneo-architecture}. The key innovations introduced aim to increase $\tau$, the number of tokens committed per speculative round, without increasing the speculative overhead $t_\text{draft}+t_\text{verify}$ (Section \ref{sec:background}). Each innovation addresses a limitation identified in Section \ref{sec:relatedwork}: MoFA removes \textit{residual mixing} from feature encoding (Section \ref{sec:mofa}), the fused causal head improves the expressivity of existing heads (Section \ref{sec:causal-head}), rejection memory lets the drafter exploit rejected proposals (Section \ref{sec:rejection-memory}), DNeo cross-attention removes redundant computation from the backbone (Section \ref{sec:prefix-only}), and the DNeo feature selector encodes the most useful target layers (Section \ref{sec:feature-protocol}).



\subsection{Feature Encoding}
\label{sec:mofa}

Existing literature has not yet isolated which target hidden states are most important for parallel drafters, nor adequately explored how different encoding operations impact $\tau$. This section proposes Mixture-of-Features Attention (MoFA), a feature encoder that avoids the residual mixing of the standard dense encoder $F_c$ (Section \ref{sec:relatedwork}), yields superior $\tau$, and makes the importance of each target hidden state interpretable.

\paragraph{MoFA fusion.} Each of the $c$ selected hidden states $\vec h_i^{(l_1)}, \dots, \vec h_i^{ (l_c)}$ is assigned an \textit{importance} scalar $u_1, \dots, u_c$, collectively parametrized by a softmax vector $u_1, \dots, u_c=\text{softmax}(w_1, \dots, w_c)$ of scalar weights. Target hidden states are weighted by importance to produce the fused feature 
\begin{align}
    \vec f_i=\text{Enc}_\text{MoFA}(x_i)&\triangleq\text{softmax}(w_{1}, \dots, w_{c}) \cdot (\vec h_i^{(l_1)}, \dots, \vec h_i^{ (l_c)}),\\
    &=u_1\vec h_i^{(l_1)}+u_2\vec h_i^{(l_2)}+\dots + u_c\vec h_i^{(l_c)},
\end{align}

defining the differentiable encoding operation $\text{Enc}_\text{MoFA}(\cdot)$, whose learned weights we call the \textit{MoFA gate}. Two properties motivate this form. First, each $u_k$ scales an entire hidden state, so every coordinate of $\vec f_i$ combines only the same coordinate across target layers: MoFA performs no residual mixing by construction (Section \ref{sec:mofa_res}). Second, the softmax turns the weights into a distribution over target layers, an attention over the depth of the target model that gives MoFA its name, so each learned scalar $u_k$ can be read directly as the importance of layer $l_k$; Section \ref{sec:feature-protocol} uses these importances to select layers. The fused feature $\vec f_i$ is re-used by all drafter layers, distinguishing MoFA from the per-layer features of DFlare \citep{zhang2026dflare} and keeping the encoder minimal. Importantly, MoFA produces consistent increases in $\tau$ relative to the standard dense feature encoder ($+16\%$, Section \ref{sec:mofa_res}) while using $\sim0.05\%$ as many encoding parameters (Section \ref{sec:results}).

\textbf{Practical considerations.} MoFA assumes that the hidden state dimensions of the draft and target model are equal, $d(\text{drafter})=d(\text{target})=d_h$, since it passes target coordinates to the drafter unchanged; this assumption is consistent with the parallel drafters of frontier open LLMs \citep{deepseekai2026deepseekv41flashpushinglimitskv, coreteam2026mimov26scalingreinforcementlearning}. However, much like the dense feature encoder, MoFA can be extended to the general case where $d(\text{drafter})\neq d(\text{target})$ by setting the input dimension of the drafter's key-value matrices equal to $d(\text{target})$. A faithful reproduction  of \textit{structure-preserving encoding} (i.e., MoFA, DFlare) requires that the model hidden dimensions are set equal to one another \citep{zhang2026dflare}.

\subsection{DNeo Causal Head}
\label{sec:causal-head}

Recall that a causal head corrects the backbone logits $\vec z_j$ left to right, conditioning each draft position on the token sampled before it, to counter suffix decay (Section \ref{sec:relatedwork}). The DSpark causal head $W_r$ \citep{cheng2026DSpark} is limited in expressivity, conditioning solely on the one-hot draft predecessor $\mathbf{1}[\hat x_{j-1}]$ with a low rank linear adjustment. Instead, the score head of DFlash-2 \citep{inco2026dflash2}, which we denote DF2, computes more expressive logit adjustments 
\begin{equation}  
   \underset{\text{logit correction}\; (1\times 1)}{\underbrace{\Delta \vec z_j(\hat x_{j} \mid \hat x_{j-1})}} =\text{DF2}(\hat x_j, \hat x_{j-1})\triangleq [\underset{(1\times r)}{\underbrace{W^V\mathbf{1}[\hat x_{j-1}]\odot W^K\;\vec h^{(\ell_m)}_{j}}}]\; \cdot\;  \underset{(1 \times r)}{\underbrace{W^Q\hat x_{j}}}
\end{equation}
using attention matrices $W^Q, W^V \in {\cal R}^{r\times|{\cal V}|}$, $W^K \in {\cal R}^{d_h \times r}$, and the hidden state $\vec h^{(\ell_m)}_j \in {\cal R}^{d_h}$ of the last draft layer $\ell_m$, which carries the context of position $j$. The logit correction $\Delta \vec z_j(\hat x_{j} \mid \hat x_{j-1}) \in {\cal R}$ is added to the logit vector $\vec z_j$ indexed by proposal $\hat x_j$, parallelized over all draft positions. However, the DF2 head is applied only to the top-k ($k=16$) logits across the draft, preventing the head from making global vocabulary adjustments.

Instead, DNeo uses the DF2 head architecture to improve the expressivity of the DSpark causal head $W_r$, by combining the outputs of the two heads in their shared rank-$r$ latent space before up-projecting to the full vocabulary:
\begin{equation}
\begin{split}
\text{DNeo}_\text{head}(\hat x_{j-1})\triangleq{}
\underset{(r \times |{\cal V}|)}{\underbrace{W_\text{up}}}\!\Bigl(
\alpha_j\cdot\underbrace{\bigl[W^V(\mathbf{1}[\hat x_{j-1}])\odot W^K\;\vec h_j^{(\ell_{m})}\bigr]\;\cdot\;W^Q[ {\cal V}]}_{\text{Dense DF2 Head} \;(1 \times r)}
\;+ \;\beta_j\cdot\underset{{\text{DSpark Head}\; (|{\cal V}|\times r)}}{\underbrace{W_{\text{down}}}(\mathbf{1}[x_{j-1}]})
\;\Bigr),
\end{split}
\end{equation}
with position-dependent learned scalar weights $\alpha_j, \beta_j \in {\cal R}$, and $W^Q[{\cal V}] \in {\cal R}^{|{\cal V}|\times r}$ denoting the low-rank query embeddings of the full vocabulary. Each element of this design addresses a limitation of one head. The DNeo head evaluates the DFlash-2 operation against the full vocabulary ${\cal V}$, rather than the top-$k$ candidates, so that its context-aware correction becomes global: $\Delta \vec z_{j}(\cdot \mid \hat x_{j-1})=\text{DNeo}_\text{head}(\hat x_{j-1}) \in {\cal R}^{|{\cal V}|}$. The latent fusion lets this context-aware term enrich the low-rank code of the DSpark head, which $W_\text{up}$ then maps to the full vocabulary, and the weights $\alpha_j, \beta_j$ let the balance between the two terms vary along the draft block. The DNeo head defines the corrected posterior
\begin{equation}
    q_{j}^{\text{DNeo}}(\hat x_{j} \mid \hat x_{j-1}, \cdot) = \text{softmax}(\vec z_{j}+\text{DNeo}_\text{head}(\hat x_{j-1})),
\end{equation}

and applies the correction left-to-right throughout the draft block following DSpark \citep{cheng2026DSpark}, so that each sampled token remains consistent with its predecessor (Section \ref{sec:relatedwork}).



\begin{figure}[t]
    \centering
    \includegraphics[width=0.999\linewidth]{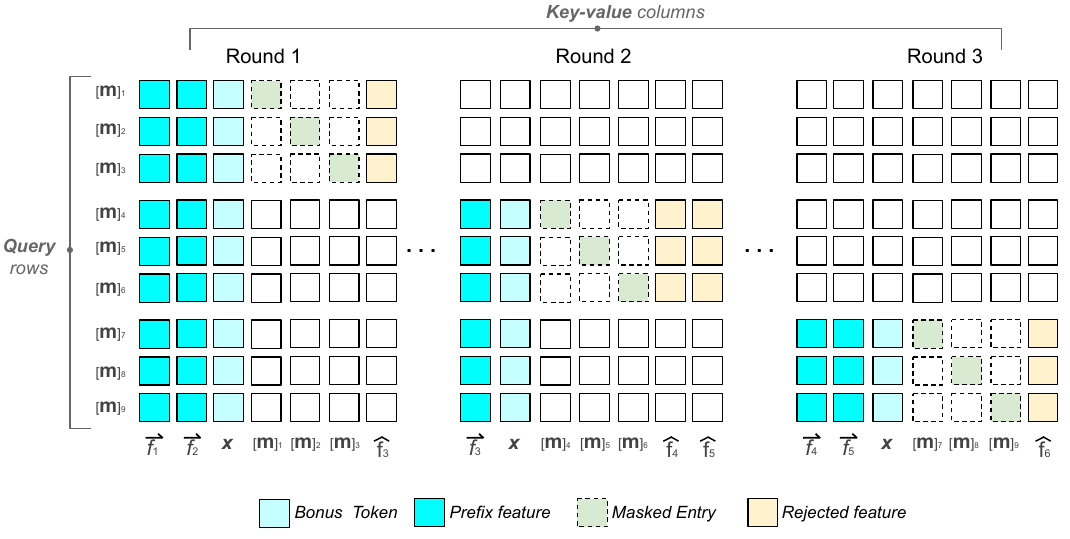}
    \caption{\textbf{DNeo Attention mask.} Mask assumes block-size $b=3$ across three drafting rounds. DNeo omits bidirectional self-attention over the masked block, a design we call prefix-only attention (Section \ref{sec:prefix-only}), and performs cross-attention to the rejected states $({\color{red} {\hat f}_3, \dots,  {\hat f}_6})$. he number of rejection features per round varies based on the prior round’s acceptance length.}
    \label{fig:dneomask}
\end{figure}

\subsection{Rejection Memory}
\label{sec:rejection-memory} 

State-of-the-art parallel drafters utilize encoded context features $\vec f_i$ while ignoring features from rejected draft tokens \citep{chen2026dflash, cheng2026DSpark, inco2026dflash2, zhang2026dflare}. However, rejected tokens are a natural source of additional information for subsequent rounds. Recall that the target model verifies the entire draft in a single forward pass (Section \ref{sec:background}): its hidden states at the rejected positions are therefore already computed, and these positions overlap those that the next round must draft.

DNeo exploits this information with \textit{rejection memory}, a novel and lightweight mechanism with two steps: \textit{rejection encoding}, which encodes the target states over the rejected postfix, and \textit{auxiliary attention}, which adds cross-attention connections from the masked block to these encoded rejections, consequently enabling the drafter to condition on recent rejection events.

\textbf{Rejection encoding.} At speculative round $t$, the draft $\hat x_{1}, \dots, \hat x_{b}$ is verified, yielding an accepted prefix ${\color{olive} (x_{1}, \dots, x_{j})}$ and a rejected postfix ${\color{red}({\hat{x}}_{j+1}, \dots, {\hat x}_{b})}$. Using the same MoFA encoder for both (Section \ref{sec:mofa}), context features are computed from the accepted prefix:
\begin{equation}
    \underset{\text{Context features}}{\underbrace{\color{olive}\vec f_{1}, \dots, \vec f_{j}}}=\text{Enc}_\text{MoFA}({\color{olive}x_{1}, \dots, x_{j}}),
\end{equation}
and \textit{rejection features} are computed from the full rejected postfix:
\begin{equation}
   \underset{\text{Rejection features}}{\underbrace{\color{red}({\hat f}_{j+1}, \dots, {\hat f}_{b})}}=\text{Enc}_\text{MoFA}({\color{red}{\hat x}_{j+1}, \dots, {\hat x}_{b}}),
\end{equation}
embedding rich information about the rejected postfix. Sharing the encoder places rejection features in the same space as context features, so the drafter can attend to both with its existing key-value projections and without additional encoding parameters.

\textbf{Auxiliary attention.} Rejection features are appended after the masked block, which allows the drafter to distinguish them from prefix features. Auxiliary cross-attention connections are added between masked positions and rejection features (see Fig. \ref{fig:dneomask} for the DNeo attention mask). The input $X_{t+1}$ to the draft model at the next round $t+1$ is:
\begin{equation}
    X_{t+1}=[\underset{\text{Context features}}{\underbrace{\color{olive}\vec f_{1}, \dots, \vec f_j}},\; x_{\text{bonus}} \;\circ \; \underset{\text{Masked block}}{\underbrace{{[\mathbf{m}]}_1, \dots, {[\mathbf{m}]}_b}},\;\;\circ\underset{\text{Rejection features (step t)}}{\underbrace{\color{red}({\hat f}_{j+1}, \dots, {\hat f}_{b})}}],
\end{equation}

where rejection features act as auxiliary context: they supply key-value states, but do not yield predictions.

Auxiliary attention enables information to pass through speculative rounds without directly altering the masked block, distinguishing rejection memory from \textit{self-conditioning} approaches \citep{diffusiongemmateam2026diffusiongemmatechnicalreport, strudel2022selfconditionedembeddingdiffusiontext, cardei2026simpleselfconditioningadaptationmasked, zhang2026dflowenablingverifierinformation}, which fuse previous predictions into the drafter's inputs. Keeping these two representations distinct preserves the role of the masked embeddings, and lets each masked position retrieve only the rejected features relevant to it (Section \ref{sec:rejmem_res}).

\textbf{Practical considerations.} Rejection memory introduces serial dependencies between speculative rounds, since $X_{t+1}$ depends on the outcome of round $t$. To avoid train-test mismatches, the drafter must therefore be trained on speculative roll-outs, through on-policy distillation \citep{lei2026draftopdonpolicydistillationspeculative, agarwal2024onpolicydistillationlanguagemodels} (Section \ref{sec:edesign}). In practice, parallel drafter training typically occurs online, alongside LLM post-training, making DNeo compatible with frontier training pipelines \citep{deepseekai2026deepseekv41flashpushinglimitskv}.

\subsection{Prefix-Only Attention}
\label{sec:prefix-only}

Parallel drafters such as DFlash perform bidirectional self-attention over the masked block (Section \ref{sec:background}), so that draft positions can coordinate their predictions. DNeo removes this self-attention: masked positions attend only to context and rejection features (Fig. \ref{fig:dneomask}), a design we call \textit{prefix-only attention}. Two observations motivate this choice. First, the dependencies between draft positions that matter most are local and left-to-right, and DNeo already captures them with the 2-tap convolutions of DFlash-2 (Section \ref{sec:relatedwork}) and with its causal head (Section \ref{sec:causal-head}). In our experiments, self-attention over the masked block brought no further gain in $\tau$ once these mechanisms were in place (Appendix \ref{appendix:prefix-only}). Second, removing self-attention eliminates the key and value projections over masked positions and shrinks the attention mask, which reduces the DNeo backbone latency (Appendix \ref{sec:latency}). 

\subsection{Layer Selection Protocol}  
\label{sec:feature-protocol}

Feature encoding typically isolates a subset of $c$ hidden states among the $c^{*}$ layers of the target model. By default, layer indices $i$ are selected uniformly across the target model:
\begin{equation}        
i_\text{uniform} \in (1, \lfloor2c^*/c\rfloor, \lfloor3c^*/c\rfloor, \ldots, \lfloor(c-1)c^*/c\rfloor, {c^*}),
\end{equation}
defining the \textit{uniform protocol} \citep{chen2026dflash, zhang2026dflare}, which provides broad coverage across target model depth, but does not select based on how useful different target layers may be to the drafter.

However, interpreting layer importance with the MoFA gate (Appendix \ref{appendix:target-layer-importnace}), whose weights $u_k$ measure the importance of each encoded layer (Section \ref{sec:mofa}), reveals that the later half of the target model contains the most important hidden states for drafting (Appendix \ref{appendix:target-layer-importnace}). This is consistent with the role of depth in transformers: later layers refine the residual stream toward the next-token prediction that the drafter must reproduce, while the first layer preserves the identity of each token. DNeo therefore proposes an adjusted protocol that keeps the first layer and draws the remaining hidden state indices $i$ from the second half of the target model:
\begin{equation}\label{eq:dneoprotocol}
   i_\text{DNeo} \in (1, \lfloor c^*/2\rfloor+\lfloor2c^*/2c\rfloor, \lfloor c^*/2\rfloor+\lfloor3c^*/2c\rfloor, \ldots, {c^*}),
\end{equation}
defining the \textit{DNeo protocol}. The DNeo protocol ultimately improves $\tau$ over the uniform protocol (Section \ref{sec:protocol_res}).

\section{Results}
\label{sec:results}

We now evaluate whether the design choices of Section \ref{sec:architecture} translate into faster lossless generation: Section \ref{sec:speed} measures speedups at batch size 1, Section \ref{sec:thruput} measures throughput under concurrency, and Section \ref{sec:tpf} isolates the contribution of each component to $\tau$.





\subsection{Experimental Design}
\label{sec:edesign}

These experiments train several SoTA parallel draft architectures \citep{chen2026dflash, inco2026dflash2, cheng2026DSpark} with fixed training conditions (i.e., data, hyperparameters, loss function, target model) for two durations: 100,000 rollouts, which we call \textit{experimental training} and use for ablations, and 1,600,000 rollouts, which we call \textit{full training} and use for the main results, with learning rate warmed linearly ($0\to 1e^{-5}$) over $20,000$ rollouts, then linearly decayed ($1e^{-5} \to 0$) over the remaining rollouts \citep{bergsma2025straightzerolinearlydecaying}. Because all parallel drafters share the same data, loss, target model, and training budget, differences in $\tau$ and speed isolate the effect of architecture.

\textbf{Training setup.} Prompts are sampled from the Llama-Nemotron post-training dataset \citep{yang2025qwen3technicalreport} across broad categories (see Appendix \ref{sec:appendix}), with four rollouts per prompt for on-policy distillation \citep{lei2026draftopdonpolicydistillationspeculative} sampled in \textit{thinking} mode. The target model is Qwen3-4B \citep{yang2025qwen3technicalreport} consistent with existing speculative decoding work \citep{chen2026dflash, cheng2026DSpark, zhang2026dflare}, with roll-outs capped at 1024 tokens.

The DSpark \citep{cheng2026DSpark} loss function defines the prediction error (${\cal L}_{\text{pred-err}}(\theta)$) during training:
\begin{equation}
    {\cal L}_{\text{pred-err}}(\theta) \triangleq \sum_{j=1}^{j=b} \text{exp}(-\frac{j-1}{b})[\lambda_{\text{L1}}\cdot\text{L1}(q_\theta(x_j), p(x_j)) + \lambda_{\text{ce}} \cdot\text{CELoss}(q_\theta(x_j), p(x_j))],
\end{equation}
for L1 loss \citep{huber1964robust}
$ \text{L1}(q_\theta(x_j),p(x_j))=\|q_\theta(x_j)-p(x_j)\|_1,$ and, cross-entropy loss \citep{goodfellow2016deep} $\text{CELoss}(q_\theta(x_j),p(x_j))=-q_\theta(x_j)\log(p(x_j))$, scaled by $\lambda_{\text{L1}}=0.9, \lambda_{\text{ce}}=0.1$, to train drafter parameters $\theta$ on the target continuation ($x_1, \dots, x_b$). (see Appendix \ref{sec:appendix} for more details).


\textbf{Inference setup.} Inference speeds are measured on 1xH100, served with SGLang. Main results use batch-size=1, (ablated in Table \ref{tab:conc}), and block-size $b=16$. Speedup is measured relative to decode-only AR speed, consistent with present literature \citep{chen2026dflash}. The evaluation suite consists of several common benchmarks (GSM8K, Math500, HumanEval, MBPP, GPQA, and MTBench) each prompt and token matched to ensure equal weighting (see Appendix \ref{sec:appendix}) \citep{cobbe2021trainingverifierssolvemath, wang2025evaluationllmsmathematicalproblem, chen2021evaluatinglargelanguagemodels, austin2021programsynthesislargelanguage, rein2023gpqagraduatelevelgoogleproofqa, Bai_2024}. The test suite remains consistent with existing drafter evaluations \citep{cheng2026DSpark, chen2026dflash}.

\begin{table}[h]
    \centering
    \small
    \setlength{\tabcolsep}{3.5pt}
    \renewcommand{\arraystretch}{1.14}
    
    \resizebox{\linewidth}{!}{%
    \begin{tabular}{@{}ll*{6}{cc}@{}}
        \toprule
        Model & Method
        & \multicolumn{4}{c}{MATH}
        & \multicolumn{4}{c}{CODE}
        & \multicolumn{2}{c}{KNOWLEDGE}
        & \multicolumn{2}{c}{CHAT} \\
        \cmidrule(lr){3-6}\cmidrule(lr){7-10}
        \cmidrule(lr){11-12}\cmidrule(lr){13-14}
        & & \multicolumn{2}{c}{GSM8K}
        & \multicolumn{2}{c}{MATH-500}
        & \multicolumn{2}{c}{HumanEval}
        & \multicolumn{2}{c}{MBPP}
        & \multicolumn{2}{c}{GPQA}
        & \multicolumn{2}{c}{MT-Bench} \\
        \midrule
        \multicolumn{2}{l}{\textbf{Temperature = 1}}
        & Speedup & $\tau$ & Speedup & $\tau$ & Speedup & $\tau$
        & Speedup & $\tau$ & Speedup & $\tau$ & Speedup & $\tau$ \\
        \midrule
        \multirow{5}{*}{Q3-4B} & EAGLE-3$^\dagger$ \textcolor{gray}{(16)}
        & $1.95\,\times$ & $4.00$ & $2.22\,\times$ & $4.57$ & $1.92\,\times$ & $3.93$ & $1.85\,\times$ & $3.80$ & $1.65\,\times$ & $3.39$ & $1.32\,\times$ & $2.72$ \\
        & DFlash-2 \textcolor{gray}{(16)}
        & $2.65\,\times$ & $3.66$ & $2.93\,\times$ & $4.05$ & $2.56\,\times$ & $3.53$ & $2.60\,\times$ & $3.60$ & $2.06\,\times$ & $2.85$ & $1.93\,\times$ & $2.68$ \\
        & DFlash \textcolor{gray}{(16)}
        & $2.65\,\times$ & $3.55$ & $2.88\,\times$ & $3.85$ & $2.62\,\times$ & $3.52$ & $2.64\,\times$ & $3.54$ & $2.00\,\times$ & $2.68$ & $1.90\,\times$ & $2.55$ \\
        & DSpark \textcolor{gray}{(16)}
        & $2.59\,\times$ & $3.87$ & $2.92\,\times$ & $4.36$ & $2.58\,\times$ & $3.85$ & $2.65\,\times$ & $3.96$ & $2.01\,\times$ & $3.00$ & $1.87\,\times$ & $2.80$ \\
        & \textbf{{\color{brown}DNeo (ours)}} \textcolor{gray}{(16)}
        & ${\color{brown}\mathbf{3.20}\,\times}$ & $({\color{brown}4.84})$ & ${\color{brown}\mathbf{3.55}\,\times}$ & $({\color{brown}5.37})$ & ${\color{brown}\mathbf{3.09}\,\times}$ & $({\color{brown}4.67})$ & ${\color{brown}\mathbf{3.08}\,\times}$ & $({\color{brown}4.66})$ & ${\color{brown}\mathbf{2.43}\,\times}$ & $({\color{brown}3.67})$ & ${\color{brown}\mathbf{2.25}\,\times}$ & $({\color{brown}3.41})$ \\
        \midrule
        \multicolumn{2}{l}{\textbf{Temperature = 0}}
        & Speedup & $\tau$ & Speedup & $\tau$ & Speedup & $\tau$
        & Speedup & $\tau$ & Speedup & $\tau$ & Speedup & $\tau$ \\
        \midrule
        \multirow{5}{*}{Q3-4B} & EAGLE-3$^\dagger$   \textcolor{gray}{(16)}
        & $2.14\,\times$ & $({4.47})$ & $2.51\,\times$ & $({5.22})$ & $2.15\,\times$ & $({4.47})$ & $2.03\,\times$ & $4.24$ & $1.85\,\times$ & $({3.86})$ & $1.43\,\times$ & $2.98$ \\
        & DFlash-2 \textcolor{gray}{(16)}
        & $2.54\,\times$ & $3.49$ & $2.85\,\times$ & $3.93$ & $2.51\,\times$ & $3.46$ & $2.53\,\times$ & $3.48$ & $1.95\,\times$ & $2.69$ & $1.88\,\times$ & $2.59$ \\
        & DFlash \textcolor{gray}{(16)}
        & $2.82\,\times$ & $3.75$ & $3.16\,\times$ & $4.19$ & $2.76\,\times$ & $3.67$ & $2.78\,\times$ & $3.70$ & ${\mathbf{2.24}\,\times}$ & $2.97$ & ${\mathbf{2.08}\,\times}$ & $2.76$ \\
        & DSpark \textcolor{gray}{(16)}
        & $2.43\,\times$ & $3.66$ & $2.76\,\times$ & $4.15$ & $2.45\,\times$ & $3.70$ & $2.47\,\times$ & $3.71$ & $1.82\,\times$ & $2.74$ & $1.75\,\times$ & $2.64$ \\
        & \textbf{{\color{brown}DNeo (ours)}} \textcolor{gray}{(16)}
        & ${\color{brown}\mathbf{2.87}\,\times}$ & $4.42$ & ${\color{brown}\mathbf{3.19}\,\times}$ & $4.91$ & ${\color{brown}\mathbf{2.82}\,\times}$ & $4.35$ & ${\color{brown}\mathbf{2.81}\,\times}$ & $({\color{brown}4.32})$ & $2.13\,\times$ & $3.27$ & $2.01\,\times$ & $({\color{brown}3.09})$ \\
        \bottomrule
    \end{tabular}%
    }    
    \caption{\textbf{Main speculative decoding results}. Bold marks the best speedup among the listed methods; parentheses mark the best TPF ($\tau$). All models are trained with the \textit{full training} setup (Section \ref{sec:edesign}), evaluated under identical conditions. ($^\dagger{}$EAGLE-3 uses official trained weights for Qwen3-4B \citep{yang2025qwen3technicalreport}). \label{table:speed-up}}
\end{table}

\subsection{Decoding Speed}\label{sec:speed}

Table \ref{table:speed-up} lists our main results, where each architecture was trained to draft (distillation) for Qwen3-4B. Results show that DNeo dominates SoTA parallel drafters in both draft-quality ($\tau$) and generation speed (\textit{speedup}). DNeo is up-to $+20.7\%$ faster than the next-best method (DFlash-2), and is always the fastest method at Temperature=1 ({see bold values}). DNeo further achieves higher $\tau$ than SoTA parallel drafters (up-to $+25.1\%\;\tau$), while exceeding the draft quality of the SoTA AR drafter EAGLE-3 at Temperature=1 (up-to $+21.0\%$ $\tau$). Notably, DNeo achieves this $\tau$ increase solely from architectural improvement, not relying on draft trees \citep{hu2026jetspecbreakingscalingceiling, li2026eagle}.

\textbf{Why DNeo is faster.} Speedup is the product of $\tau$, which DNeo raises through the components analyzed in Section \ref{sec:tpf}, and the cost of each speculative round. EAGLE-3 exceeds the $\tau$ of DNeo on four benchmarks at Temperature=0, yet DNeo is $1.15\times$ to $1.34\times$ faster there, because EAGLE-3 drafts autoregressively while DNeo drafts the entire block in one forward pass. Among parallel drafters, the left-to-right causal head makes a DNeo round costlier than a DFlash round, so DNeo is fastest whenever its $\tau$ advantage outweighs this overhead: on every benchmark at Temperature=1 ($+32\%$ to $+40\%\;\tau$ over DFlash) and on math and code at Temperature=0 ($+17\%$ to $+19\%$), but not on GPQA and MT-Bench at Temperature=0 ($+10\%$ to $+12\%$). The advantage peaks under sampling because a parallel drafter averages each prediction over all plausible predecessors, is less problematic for greedy decoding than for sampling; causal heads correct exactly this mismatch (Section \ref{sec:relatedwork}), which is also why DSpark and DFlash-2 improve on the $\tau$ of DFlash mainly at Temperature=1.


\subsection{Higher Throughput}\label{sec:thruput}

The advantage of DNeo in committed tokens per forward ($\tau$) results in improved throughput, motivating the evaluation of DNeo's total generated tokens per second under different concurrency values (batch sizes, $B$; Table. \ref{tab:conc}). DNeo exhibits higher throughput than other speculative methods in all settings on the evaluation set (Section \ref{sec:edesign}). DNeo also achieves up-to $+58\%$ higher throughput (at B=32) than DFlash-2 (Transformers serving). The gap widens under Transformers serving because DNeo's throughput advantage there exceeds its advantage in $\tau$: its rounds are also cheaper than those of the baselines, consistent with its leaner prefix-only backbone (Section \ref{sec:prefix-only}). This throughput gap can be attributed to the higher $\tau$ DNeo achieves, and the drafter's competitive speculative latency (Appendix \ref{sec:latency}). A high $\tau$ also keeps speculation profitable as concurrency grows: at small $B$, decoding is memory-bound and verifying a draft block costs little more than generating one token, but as $B$ grows, verification approaches the compute limit of the GPU and every rejected draft token wastes compute that other requests could use. All speculative methods thus lose ground to AR, and DNeo, which commits the most tokens per round, wastes the least. The net result is that DNeo dominates the throughput-speed Pareto (Fig. \ref{fig:motivation}), being the only drafter to maintain throughput parity with the AR baseline at batch size 64 (within $0.1\%$), while the other drafters fall $10\%$ to $18\%$ below it.

\begin{table*}[h]
\centering
\small
\setlength{\tabcolsep}{2pt}
\renewcommand{\arraystretch}{1.22}
\begin{tabular*}{\linewidth}{@{\extracolsep{\fill}}lrrrrr@{\hspace{10pt}}rrrrr@{}}
\toprule
& \multicolumn{5}{c}{\textbf{SGLang}} & \multicolumn{5}{c}{\textbf{Transformers}} \\
\cmidrule(lr){2-6}\cmidrule(l){7-11}
Method & $B=1$ & $B=4$ & $B=16$ & $B=32$ & $B=64$
       & $B=1$ & $B=4$ & $B=16$ & $B=32$ & $B=64$ \\
\midrule
AR & 234 & 890 & 3,310 & 5,965 & \textbf{10,335}
   & 124 & 364 & 733 & 900 & 1,023 \\
EAGLE-3 & 442 & 1,483 & 4,409 & 6,576 & 8,446
        & 174 & 547 & 1,124 & 1,352 & 1,457 \\
DFlash & 589 & 1,964 & 5,770 & 7,797 & 8,834
       & 269 & 762 & 1,223 & 1,388 & 1,462 \\
DFlash-2 & 590 & 2,023 & 5,781 & 7,971 & 9,318
         & 272 & 743 & 1,303 & 1,475 & 1,511 \\
DSpark & 586 & 1,918 & 5,485 & 7,435 & 9,077
       & 261 & 706 & 1,278 & 1,404 & 1,460 \\
\textbf{{\color{brown}DNeo}} ({\color{brown}\textbf{ours}}) & {\color{brown}{\color{brown}\textbf{705}}} & {\color{brown}\textbf{2,247}} &{\color{brown} \textbf{6,580}} & {\color{brown}\textbf{9,058}} & {10,324}
            & {\color{brown}\textbf{404}} & {\color{brown}\textbf{1,101} }&{ \color{brown}\textbf{2,033}} & {\color{brown}\textbf{2,201} }&{ \color{brown}\textbf{2,269}} \\
\bottomrule
\end{tabular*}
\caption{\textbf{Decode throughput}. (tokens/s) on 1xH100. verifier and drafter temperature 1, $b=16$. Bold marks the best method per column. All entries use the same six benchmarks. $B$ denotes submitted concurrency.}
\label{tab:dneo_concurrency_decode_tps_grid}
\label{tab:conc}
\end{table*}

\subsection{Architectural Ablations}\label{sec:tpf}

The architectural components introduced (i.e., rejection memory, MoFA, DNeo head) are ablated to determine their influence on drafter $\tau$, along with innovations from existing literature (i.e., DFlash-2 convolutions, DFlare encoding). Notably, MoFA, DFlash-2 convolutions, rejection memory, DNeo-head, all result in draft quality ($\tau$) improvements (Table \ref{tab:dneo_independent_component_ablation}); since each is removed independently, none is made redundant by the others. MoFA improves draft quality the most ($+0.651\;  \tau$), due to its target feature structure preserving properties evaluated in Section \ref{sec:mofa_res}. The convolutions rank second ($+0.331\;  \tau$), serving as the only channel between neighboring draft positions inside the backbone. Importantly, the proposed innovations, rejection memory ($+0.182\;  \tau$) and the fused DNeo head ($+0.217\;  \tau$), also contribute to the recipe, for the reasons analyzed in Sections \ref{sec:head_res} and \ref{sec:rejmem_res}. All configurations use the DNeo protocol, which is evaluated in detail in section \ref{sec:protocol_res}.

\begin{table}[h]
\centering
\caption{\textbf{Subtractive component ablation}. Each component is removed independently from DNeo, not cumulatively.}
\vspace{+5pt}
\label{tab:dneo_independent_component_ablation}
\begin{tabular}{lrr}
\toprule
Architecture & Train $\tau$ (\textit{100,000 prompts}) $\uparrow$ & $\Delta\tau$ \\
\midrule
\textbf{DNeo} ($c=7$) & \textbf{3.112} & --- \\
\quad $-$ MoFA (replace with $F_c$)
    & 2.461 & $-\mathbf{0.651}$ \\
\quad $-$ DFlash-2 convolutions
    & 2.780 & $-0.331$ \\
\quad $-$ Rejection memory
    & 2.930 & $-0.182$ \\
\quad $-$ DNeo head (replace with DSpark, $r=256$)
    & 2.895 & $-0.217$ \\
\bottomrule
\end{tabular}
\end{table}

\subsubsection{MoFA}\label{sec:mofa_res}

This work discovers a novel effect to explain why our proposed MoFA improves $\tau$ over standard dense encoding ($F_c$):
\begin{quote}
   \textit{Dense encoding mixes target representations across the residual dimension $d_h$, an effect we call residual mixing, damaging contextual structure and reducing $\tau$.}
\end{quote}
\textbf{Residual mixing.} Firstly, note that each feature scalar $f_j \in \vec f$ encoded by $F_c$ is the sum of $c$ vector dot-products between $F_c$ and the $c$ target hidden states:
\begin{equation}
    f_j=\sum_{i=1}^{i=c}[{\color{red}w_{1,i}},{\color{red}w_{2,i}}, \dots, {\color{teal}w_{j,i}}, \dots, {\color{red}w_{{d_h},i}}] \cdot \vec h^{(l_i)}
\end{equation}
for scalar matrix entries $w_{j,k} \in F_c$ (Fig. \ref{fig:resid-mix}). The red entries above mix information from scalars $h_{k,i} \in h^{(l_i)}$ into $f_j$ where $j\neq k$: this is residual mixing. These entries can be masked from $F_c$ (set $w_{j,k}=0$), yielding a \textit{restricted} encoding matrix that is otherwise structurally identical to $F_c$, but with no residual mixing:
\begin{align}
    F^{\text{ (restricted)}}_c \triangleq 
    \begin{bmatrix}
        W^1 \\
        W^2 \\
        \dots \\
        W^c
    \end{bmatrix}\; \text{s.t.}\; W^i \in \text{Diag}({\cal R}^{d_h \times d_h}), \forall i \in 1 \ldots c,
\end{align}
which is simply a column of $c$ diagonal matrices $W_1, \dots, W_c$. Neither the restricted matrix, nor MoFA perform residual mixing, therefore comparing the two encoders will help us determine the extent to which MoFA's advantage over $F_c$ is due to residual mixing.

\begin{wraptable}{r}{0.45\textwidth}
    \centering
    \vspace{-25pt}
    \caption{\textbf{Feature encoder ablation}. Isolating the effect of \textit{residual mixing}.}
    \vspace{3pt}
    \label{table:restrictedfc}
    \small
    \setlength{\tabcolsep}{4pt}
    \begin{tabular}{@{}lr@{}}
        \toprule
        Architecture & \shortstack[r]{Train TPF\\(50K prompts) $\uparrow$} \\
        \midrule
        DFlash-2 ($F_c$)            & 1.938 \\
        DFlash-2 (Restricted $F_c$) & 2.361 \\
        DNeo (MoFA)                & \textbf{2.526} \\
        DNeo (Restricted $F_c$)     & 2.517 \\
        \bottomrule
    \end{tabular}
    \vspace{-10pt}
\end{wraptable}

Removing residual mixing from $F_c$ (Table \ref{table:restrictedfc}) produces nearly identical $\tau$ to MoFA (within $0.3\%$), outperforming the dense encoder $F_c$ ($+0.423\;\tau$ at 50,000 roll-outs). These results isolate residual mixing as essentially the sole determinant of MoFA's advantage over $F_c$. Because $F_c$ contains $F_c^{\text{(restricted)}}$ as a special case, its deficit is not one of capacity: it must learn from data a coordinate-wise structure that MoFA and $F_c^{\text{(restricted)}}$ build in. This structure is natural for transformers, whose layers all write into one residual stream, so the hidden states of different layers share coordinates that dense projection entangles. Since MoFA yields the same draft quality as $F_c^{\text{(restricted)}}$ with fewer parameters, MoFA remains the default encoder for DNeo.

\begin{wraptable}{r}{0.45\textwidth}
    \centering
    \caption{\textbf{Causal head ablation}. Fused DNeo vs. DFlash-2, and DSpark heads (all heads use projection rank $r=256$).}
    \vspace{3pt}
    \label{table:heads}
    \small
    \setlength{\tabcolsep}{4pt}
    \begin{tabular}{@{}lr@{}}
        \toprule
        Architecture & \shortstack[r]{Train TPF\\(100K prompts) $\uparrow$} \\
        \midrule
        DSpark + DFlash-2 Head            & 2.217 \\
        DSpark + DSpark Head & 2.273 \\
        \midrule
        DSpark + \textbf{DNeo Head} (\textit{ours})                & \textbf{2.361} \\
        \bottomrule
    \end{tabular}
\end{wraptable}



\subsubsection{DNeo Head}\label{sec:head_res}

The DNeo causal head (introduced in Section \ref{sec:causal-head}) fuses the outputs of the standard low rank DSpark head ($W_r$, Section \ref{sec:relatedwork}), and the DFlash-2 score head (Section \ref{sec:causal-head}) in latent space \citep{cheng2026DSpark, inco2026dflash2}. Interestingly, the fused head (Table \ref{table:heads}) exhibits $+0.056 \;\tau$ and $+0.141 \;\tau$ over the DFlash-2 head, and DSpark head alone (respectively), highlighting the complementary inductive biases of the two heads anticipated in Section \ref{sec:causal-head}: the DSpark head adjusts the full vocabulary but sees only the previous token through a low-rank linear map, while the DFlash-2 head is context-aware but only re-ranks the top-16 candidates. The fused head applies context-aware corrections over the full vocabulary, and can therefore promote tokens that the backbone ranks outside its top candidates.

\subsubsection{Rejection Memory}\label{sec:rejmem_res}

The concurrent work DFlow \citep{zhang2026dflowenablingverifierinformation} (Section \ref{sec:relatedwork}) fuses a rejection feature ($\hat f_{j+1}$) with masked embeddings $[\mathbf{m}]_{j+1}, \dots, [\mathbf{m}]_{b}$ on the subsequent draft round using an MLP. However, the DFlow approach ignores the rightmost rejection features ($\hat f_{j+2}, \dots, \hat f_{b}$), and compresses rejection feature information into a set of fixed mask embeddings. Instead, DNeo's rejection memory uses cross-attention to enable each mask token to attend to features of all rejected tokens. As discussed in Section \ref{sec:rejection-memory}, these features are informative because the rejected positions overlap those that the next round must draft, and they come at no extra cost for the target model, which already computed their hidden states during verification.

Ablating these decisions in Table \ref{table:rej-mem} reveals that fusing rejection features with masked representation (DFlow, no MLP) produces no gain over the (memory-less) DNeo baseline ($+0.009\;\tau$), and gating the fusion with an MLP (DFlow, $d_{\textit{MLP}}=512$) achieves a small gain ($+0.030\;\tau$). Notably, combining this fusion with rejection memory (DFlow + rejection memory) produces lower $\tau$ than rejection memory alone ($-0.034\;\tau$) showing that mixing masked and rejected representations can serve as a quality bottleneck. Fusion adds one rejection feature to every masked embedding regardless of its relevance, perturbing the inputs from which the drafter predicts; in contrast, attention lets each position retrieve only the rejected features it needs.

\begin{quote}
    \textit{Rejected proposals are a valuable source of context, which are best exploited via cross-attention rather than input-fusion.}
\end{quote}

\begin{table}[h] 
\vspace{-5pt}
\centering
\caption{\textbf{Rejection attention ablation}. Gain in {$\tau$} reported relative to \textit{memory-less} DNeo baseline.}
\vspace{+10pt}
\label{tab:dneo_online_memory_ablation}
\begin{tabular}{lrr}
\toprule
Architecture & Train $\tau$ $\uparrow$ & $\Delta\tau$ \\
\midrule
\textbf{DNeo Rejection memory} (\textit{ours})
    & \textbf{3.112} & $+0.189$ \\
\midrule
\textit{DFlow} + Rejection memory (\textit{ours})
    &3.078 & $+0.155$ \\
\textit{DFlow} fusion (\textit{MLP}, $d_{\text{MLP}}=512$)
    & 2.953 & $+0.030$ \\
\textit{DFlow} fusion (\textit{no MLP})
    & 2.932 & $+0.009$ \\
\midrule
No Rejection memory/fusion (\textit{memory-less baseline})
    & 2.923 & $-$ \\    
\bottomrule
\end{tabular} \label{table:rej-mem}

\smallskip
\vspace{-10pt}
\end{table}

\subsubsection{Layer Selection Protocol}\label{sec:protocol_res}

Finally, we compare the DNeo protocol with the uniform protocol of DFlash and DFlare (Section \ref{sec:feature-protocol}) on a DSpark drafter with MoFA encoding. With the DNeo protocol, $\tau$ grows with the number of encoded target layers, from $2.481$ at $c=5$ to $2.516$ at $c=15$, whereas with the uniform protocol it decreases, from $2.467$ to $2.447$; the gap between the two protocols thus widens from $+0.014\;\tau$ to $+0.069\;\tau$ as more layers are encoded (appendix Fig. \ref{fig:dflarevsdneo}). Uniform selection draws each added layer from the full depth of the target model, so many additions come from early layers that carry little drafting signal and dilute the MoFA gate; the DNeo protocol draws every added layer from the later half, so additional layers contribute additional signal. For the same reason, MoFA over all $36$ target layers ($\tau=2.478$) does not outperform the DNeo protocol with only $5$ layers: its gate must spread its mass over many uninformative layers (Appendix \ref{appendix:dneo-protocol}).

\begin{quote}
    \textit{The DNeo protocol encodes more useful target layers, enabling better draft quality under fixed encoding conditions.}
\end{quote}

\section{Conclusion}

This paper presented DNeo, a parallel draft architecture that conditions the drafter on structure-preserving features of the target model and on the features of previously rejected proposals, and corrects its parallel predictions with a fused causal head. It showed that dense feature encoders degrade target features through residual mixing, an effect that MoFA, a lightweight mixture of target layers, removes by construction. Experiments in realistic serving settings show that the resulting gains in draft quality culminate in up to 1.3$\times$ faster generation than state-of-the-art parallel drafters. DNeo illustrates the importance of architectural research in parallel models: a rich understanding of which target features matter, and of how to encode them, translates directly into faster lossless inference.



\bibliography{references}
\bibliographystyle{iclr2027_conference}

\clearpage
\appendix
\section{Appendix}
\label{sec:appendix}

\subsection{TPF Scaling}\label{appendix:training}

\begin{figure}[h]
        \centering
    \paperfigure{0.9\linewidth}{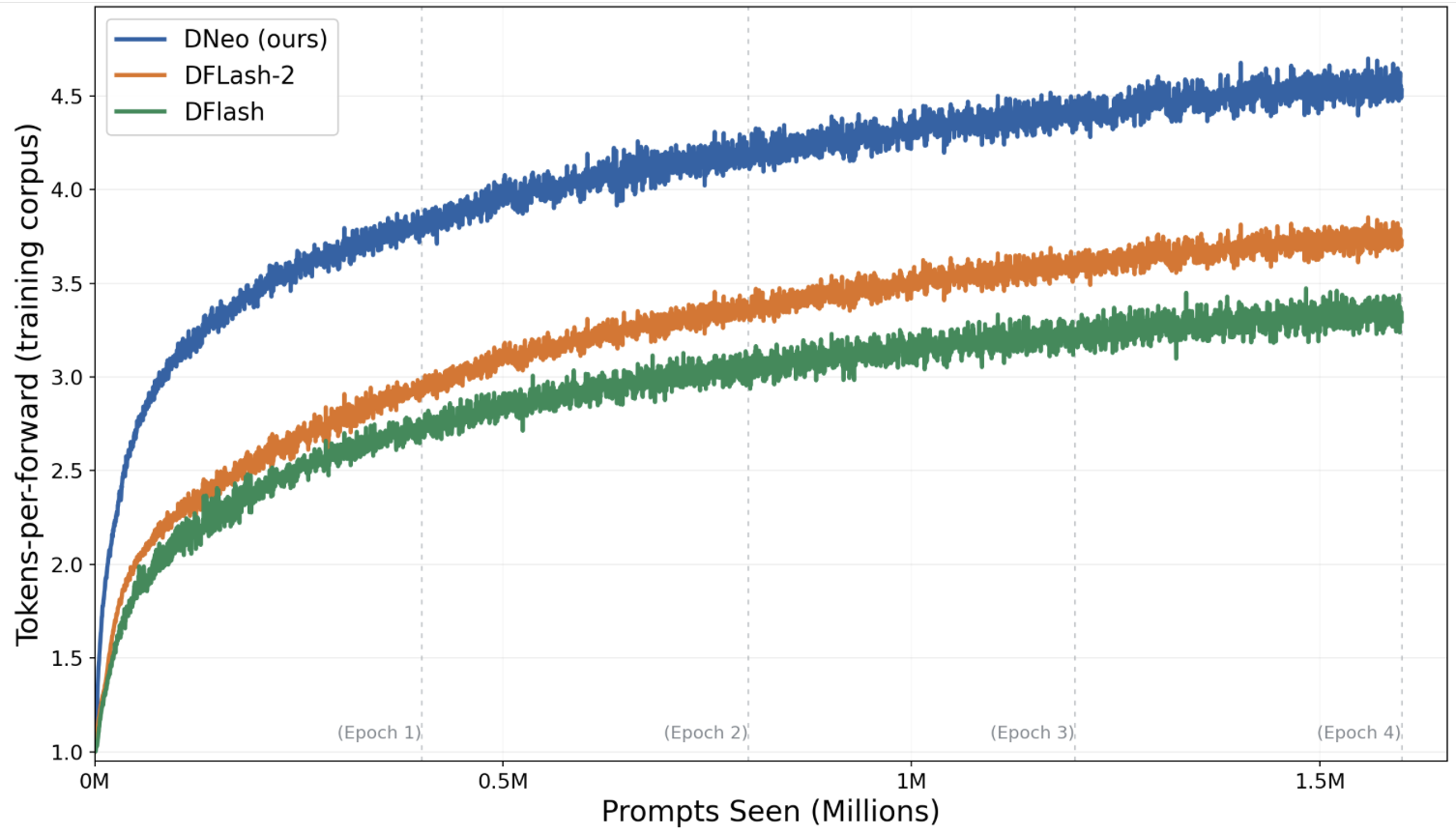}
                \caption{DNeo, DFlash and DFlash-2 train-time TPF scaling over 1.6M prompts. DNeo exhibts significant convergence advantages ($10\times$ faster and cheaper than DFlash).}
        \label{fig:training-scaling}
\end{figure}

Here we show the training curves for some of our \textit{converged} drafters on the full training set (Fig. \ref{fig:training-scaling}). The curves highlight (\textit{i}) the sustained train-time drafter quality advantage of the DNeo recipe, and (\textit{ii}) the convergence speed advantages of DNeo (up-to $10\times$ faster convergence than DFlash at matched $\tau$). We leave the analysis of parallel drafter scaling laws to future work, though stable trends appear evident. 

\subsection{Target Layer Importance}\label{appendix:target-layer-importnace}

\begin{figure}[h]
    \vspace{-0pt}
    \centering
    \paperfigure{0.9\linewidth}{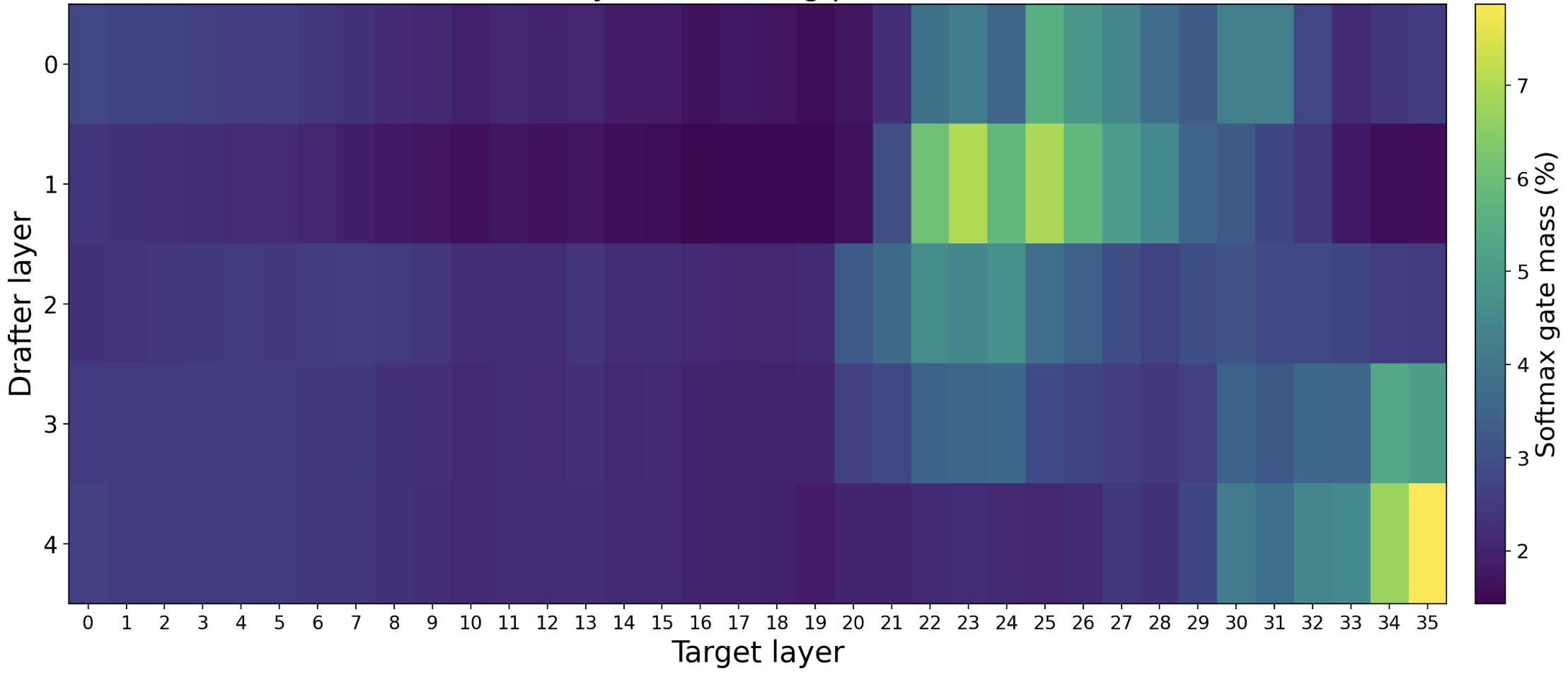}
    \vspace{-0pt}
    \caption{MoFA gates learned allocation after training (100,000 prompts). More attention mass is allocated to later target layers, a pattern that is consistent across the five trained MoFA gates (one for each draft layer).}
    \label{fig:mofa-attention}
\end{figure}

Training five MoFA gates (one for each draft layer, with $c=n$) for exploratory purposes reveals a profile reflecting relative target layer importance (MoFA profile, Fig. \ref{fig:mofa-attention}). The profile shows that the later subset of target layers (layers 20 $\to$ 35) carry noticeably more importance than the initial layers, motivating the DNeo protocol (Eq. \ref{eq:dneoprotocol}).

\subsection{Extended Ablations}\label{appendix:ablations}

Combining all architectural components additively reveals the gain in draft quality (and thus speed) that can be achieved via compounded architectural research in parallel drafters (Fig. \ref{tab:dneo_additive_ablation}). MoFA still shows the largest gain ($+0.346\; \tau$) followed by the DFlash-2 convolutions ($+0.236\; \tau$), yet relative gains are likely order dependent. The overall gain of ($+0.989\; \tau$) from DFlash to DNeo constitutes approximately a full additional generated token per forward pass, contributing to faster decoding speeds.

\begin{table}[h]
    \centering
    \caption{\textbf{Full additive architecture ablation.} Upgrading from DFlash to DNeo. Training TPF is measured at 100,000 prompts. DNeo gains almost a full token ($0.989$ tokens per forward) over DFlash.}
    \vspace{+10pt}
    \label{tab:dneo_additive_ablation}
    \begin{tabular}{lrr}
    \toprule
    Architecture & Train TPF (\textit{100,000 prompts}) $\uparrow$ & $\Delta$ TPF \\
    \midrule
    DFlash & 2.072 & --- \\
    \quad $+$ DSpark head ($r=256$) & 2.132 & $+0.060$ \\
    \quad $+$ dense MoFA conditioning ($c=n$) & 2.478 & \textbf{$+\mathbf{0.346}$} \\
    \quad $+$ DNeo MoFA Protocol ($c=5$) & 2.505 & $+0.027$ \\
    \quad $+$ per-layer MoFA routing & $2.505$ & $+0.001$ \\
    \quad $+$ dense DFlash-2 score head & $2.590$ & $+0.085$ \\
    \quad $+$ fused DNeo head & $2.646$ & $+0.056$ \\
    \quad $+$ DFlash-2 convolutions & $2.882$ & \textbf{$+0.236$} \\
    \quad $+$ DNeo prefix-only attention & $2.886$ & $+0.004$ \\
    \quad $+$ \textbf{DNeo} rejection memory & $\mathbf{3.062}$ & $+0.176$ \\
    \midrule
    \multicolumn{2}{l}{DNeo improvement over DFlash} & $\mathbf{+0.989}$ \\
    \bottomrule
    \end{tabular}
\end{table}

\subsection{Prefix Only Attention}\label{appendix:prefix-only}

DNeo cross attention omits self-attention. The motivation for this is expressed in the results here, showing that self-attention has negligible influence on drafter quality. In fact, removing self-attention results in a $+0.027\; \tau$ and $+0.004\; \tau$ in no-convolution, and convolution cases respectively.

\begin{table}[h]
\vspace{+10pt}
\centering
\caption{\textbf{Prefix only attention comparison.} Ablation of masked-token self-attention and DFlash-2 convolutions at 100,000 training prompts.}
\vspace{+10pt}
\label{tab:prefix_only_convolution_ablation}
\begin{tabular}{lrrr}
\toprule
Attention pattern
    & No convolutions
    & With convolutions
    & Convolution gain \\
\midrule
Full masked-token self-attention
    & 2.620
    & 2.855
    & $+0.236\;\tau$ \\
Prefix-only attention
    & 2.646
    & \textbf{2.859}
    & $+0.213\;\tau$ \\
\midrule
Prefix-only gain
    & $+0.027\;\tau$
    & $+0.004\;\tau$
    & --- \\
\bottomrule
\end{tabular}
\end{table}

\subsection{DNeo Protocol}\label{appendix:dneo-protocol}

Training DSpark with both the DNeo protocol, and the standard (uniform selection) protocol reveals that advantages of intelligent layer selection. In particular a stable trend emerges; as the number of utilized target features ($c$) increases, so too does the drafter quality (tokens-per-forward). The uniform selection protocol does not exhibit this trend (Section \ref{sec:protocol_res}).

\begin{figure}[h]
    \centering
    \includegraphics[width=0.75\linewidth]{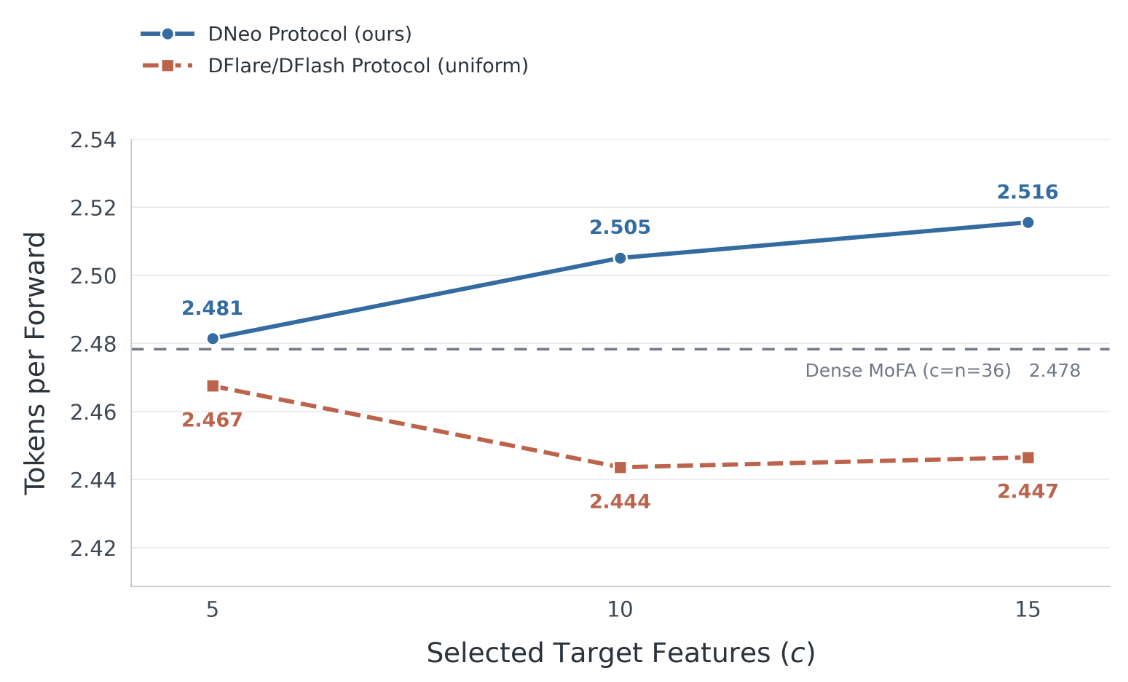}
    \caption{\textbf{DNeo protocol vs uniform selection.} Feature selectors ablated over fixed DSpark architecture, w.r.t selected layers $c$.}
    \label{fig:dflarevsdneo}
\end{figure}

\subsection{DNeo Latency}\label{sec:latency}

\begin{figure}[h]
    \centering
    \includegraphics[width=1\linewidth]{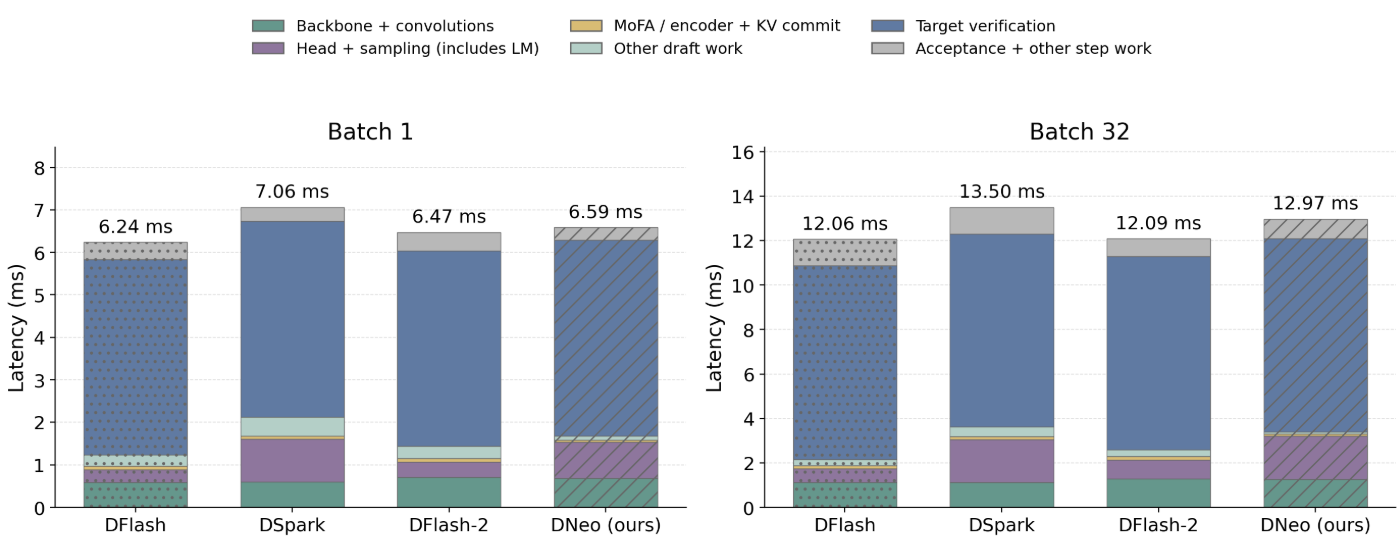}
    \caption{SGLang latency breakdown on 1xH100. block-size$=16$, precision bf$16$, for all methods, excludes prefill.}
    \label{fig:latencystack}
\end{figure}

While the draft-quality (\textit{TPF}) advantages of DNeo are clear, it is also integral to evaluate how the combined architectural innovations impact speculative latency, as a core goal of this work is to avoid substantially impacting latency ($t_{\text{draft}}+t_{\text{verify}}$). Interestingly, not only is speculative-latency not significantly increased by DNeo, but is in fact reduced against key-baselines like DFlash-2 and DSpark (Fig. \ref{fig:latencystack}). This is due to high-quality implementation, and the novel \textit{prefix-only attention}.
\clearpage

\subsection{Residual Mixing}\label{sec:resid-mix}

\begin{figure}[h]
    \centering
    \includegraphics[width=0.8\linewidth]{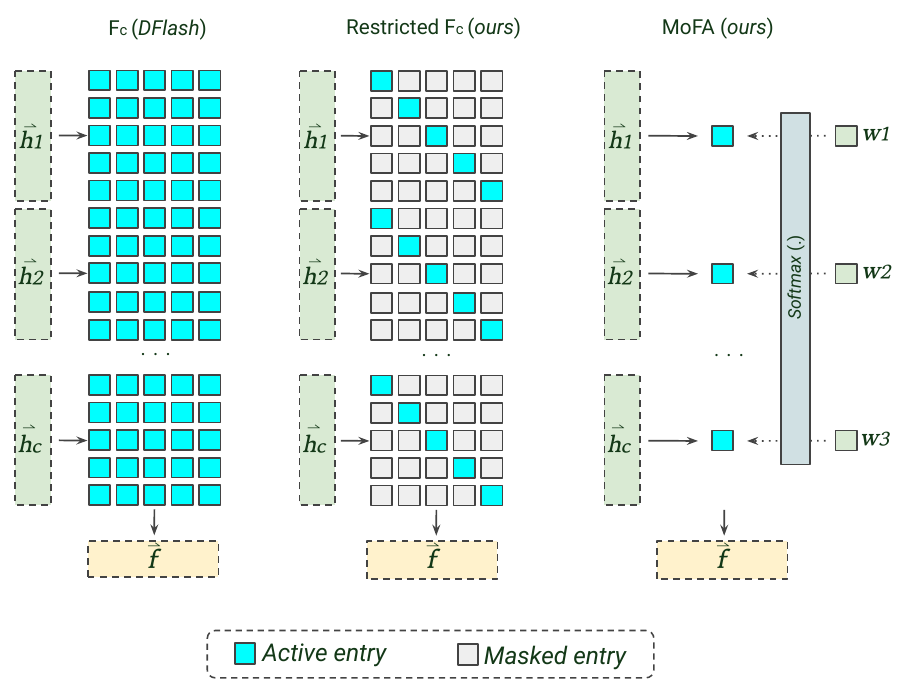}
    \caption{\textbf{Encoding matrix illustration}.}
    \label{fig:resid-mix}
\end{figure}

The restricted $F_c$ encoding matrix (Section \ref{sec:mofa_res}) masks out entries responsible for mixing across target hidden state dimensions (residual mixing). A visual illustration of this masking construction is referenced here (Fig. \ref{fig:resid-mix}). Comparatively, the MoFA encoder also does not posses entries responsible for residual mixing, amounting to simply a normalized vector of dimension $c$.

\end{document}